%% file: acl_latex.tex
\pdfoutput=1
\documentclass[11pt]{article}

\usepackage[preprint]{acl}
\usepackage{multirow}
\usepackage{times}
\usepackage{latexsym}
\usepackage{xcolor}

\usepackage[T1]{fontenc}
\usepackage[utf8]{inputenc}

\usepackage{microtype}

\usepackage{inconsolata}

\usepackage{graphicx}

\title{What Are They Filtering Out? An Experimental Benchmark of Filtering Strategies for Harm Reduction in Pretraining Datasets}

\author{Marco Antonio Stranisci \\
  University of Turin \\
  aequa-tech \\
  \texttt{marcoantonio.stranisci@unito.it} \\\And
  Christian Hardmeier \\
  IT University of Copenhagen \\
  \texttt{chrha@itu.dk } \\}

\begin{document}
\maketitle
\begin{abstract}
Data filtering strategies are a crucial component to develop safe Large Language Models (LLM), since they support the removal of harmful contents from pretraining datasets. There is a lack of research on the actual impact of these strategies on vulnerable groups to discrimination, though, and their effectiveness has not been yet systematically addressed. In this paper we present a benchmark study of data filtering strategies for harm reduction aimed at providing a systematic evaluation on these approaches. We provide an overview $55$ technical reports of English LMs and LLMs to identify the existing filtering strategies in literature and implement an experimental setting to test their impact against vulnerable groups. Our results show that the positive impact that strategies have in reducing harmful contents from documents has the side effect of increasing the underrepresentation of vulnerable groups to discrimination in datasets. 
\\
WARNING: the paper could contain racist, sexist, violent, and generally offensive contents
\end{abstract}

\input{latex/sections/1.introduction}
\input{latex/sections/2.related}
\input{latex/sections/3.overview}
\input{latex/sections/4.experiments}
\input{latex/sections/5.discussion}
\input{latex/sections/6.conclusion}

\section*{Limitation}
A first limitation of our research is the imbalance of the two knowledge bases that we use to obtain sociodemographic information about named entities. Both Wikipedia and Caligraph contain a higher number of Westerners against Post-colonial people and a higher number of men against women. This introduces a bias in the Entity Linking process that might have an impact on results. Developing more balanced datasets for the implementation of our pipeline is one of the key actions of our future work.

A second limitation regards the focus on named entities. People belonging to vulnerable groups are mentioned in documents in very different ways (e.g., demonym, pronouns, countries of origin). However, there are no systems that are trained to automatically identify these triggers of vulnerable identities. We preferred to rely on state of the art approaches to maximize the precision of our EL pipeline against the recall. The second version of our pipeline will account for these alternative ways of mentioning people with certain sociodemographic characteristics.

\section*{Ethical Consideration}
Since our work focuses on vulnerable groups to discrimination, we are aware of the risk of adopting research design biases that can have a negative influence on our categorization of people. We followed the theoretical background emerging from post-colonial and black studies, to avoid the risk of inducing stereotypical representation of vulnerable groups in our analysis. We also acknowledge that the underrepresentation of post-colonial people in existing knowledge bases can be a source of additional discrimination that will be challenged in the next iterations of our benchmarking pipeline.

\bibliography{anthology,custom}
\end{document}

%% file: latex/sections/1.introduction.tex
\section{Introduction}

The harmfulness of Large Language Models (LLM) is an open issue that gathers the attention of different sectors of our society. International bodies regulated the development of these technologies \cite{edwards2021eu}; Natural Language Processing (NLP) scholars are introducing a series of approaches \cite{touvron2023llama} to assess and mitigate their impact against vulnerable groups to discrimination. 

Even if the development cycle of a LLM encompasses several steps, in recent years most of the research focuses on the post-training stage, for which several benchmarks \cite{gehman2020realtoxicityprompts,tedeschi2024alert} have been created. Theoretical research on effective strategies to filter out harmful contents from pretraining datasets is an understudied topic, though. If compared to the amount of LLMs released in recent years, only a limited number of approaches to filtering strategies has been proposed \cite{raffel2020exploring,brown2020language}, and many o them have been implemented without considering the complex nature of bias, producing unwanted negative effects against several categories of people \cite{dodge2021documenting,xu2021detoxifying}. Few critical studies on filtering strategies \cite{luccioni2021s,longpre-etal-2024-pretrainers} have been performed so far, but none of them systematically addresses the topic. 

The aim of our research is to propose the first systematic analysis of data filtering strategies for harm reduction in pretraining dataset. Specifically, we formulate two research questions.

\textbf{RQ 1: Which filtering strategies are implemented to remove harmful contents from pretraining datasets?} We surveyed $55$ technical reports describing English LMs and LLMs to collect information about the characteristics of the existing data filtering strategies and their documentation. Through the survey we have been able to identify eight different categories of filtering strategies for harm reduction that have been proposed in literature. The survey also shows a disengagement trend in current LLM technical reports, from which emerges a general lack of awareness on their role in increasing the underrepresentation of vulnerable groups to discrimination in datasets.

\textbf{RQ 2: Which categories of people are most affected by filtering strategies?} We performed a benchmark analysis on seven data filtering strategies to evaluate if and to which extent they increase the underrepresentation of vulnerable groups to discrimination in pretraining datasets. To perform this analysis, we designed a pipeline for identifying the mentions of named entities categorized by their gender and origin. Results of our analysis show that women are the most impacted by filtering strategies and that strategies significantly differ in the contents they filter out. Choosing a specific strategies means focusing on specific sources of harm overlooking others. 

The contribution of our work is threefold: \textit{i.} we provide the first systematic survey of data filtering strategies for harm reduction; \textit{ii.} we implement a pipeline and a set of resources to benchmark filtering strategies; \textit{iii.} we test our pipeline on a pool of data filtering strategies, empirically demonstrating the negative impact of these approaches in the underrepresentation of vulnerable groups to discrimination. 

%% file: latex/sections/2.related.tex
\section{Related work}
Research on data filtering strategies for harm reduction has been propelled by the research on biases in LLMs \cite{bender2021dangers} and in pretraining datasets \cite{jo2020lessons}. However, critical studies on strategies focused on specific datasets rather than the impact of strategies themselves. \citet{luccioni2021s} analyzed the presence of Hate Speech in the Common Crawl dataset. \citet{dodge2021documenting} documented the C4 corpus \cite{raffel2020exploring}, showing that its filtering process correlates with a reduction of terms defining vulnerable groups to discrimination in datasets. \cite{longpre-etal-2024-pretrainers} tested the correlation between removing toxic contents from pretraining datasets and LLMs performance in toxicity classification, showing that filtering has a negative impact on this task. The survey of \citet{albalak2024survey} on data selection for LLMs provide a marginal and non-systematic taxonomy of strategies. Finally, the work of \cite{lucy2024aboutme} on language filtering strategies provide insightful results on the cultural biases embedded in language filters that can determine the exclusion of categories of people who speak non-standard English varieties.

Our work fills the existing gap providing a first systematic analysis of the impact of filtering strategies for harm reduction in datasets.

%% file: latex/sections/3.overview.tex
\section{An Overview of Existing Data Filtering Strategies} \label{sec:survey}
In this section we present our systematic overview of data filtering strategies for harm reduction in pretraining data as they are presented in $55$ LMs technical reports. We first introduce our methodology for the selection of relevant documents and their analysis. Then, we present a taxonomy of eight existing data filtering strategies that emerges from the analysis. Finally, we identify some trends emerging from a general overview of reports in a diachronic perspective. 

\subsection{Overview Methodology} \label{ss:survey_methodology}
To ensure a representative pool of technical reports describing LLMs, we seeded them from six leaderboards that have been chosen to include in our study different generations of LMs as well as different tasks. To ensure that small LMs were included in our survey we gathered all the technical reports of systems ranked higher than the baseline of SuperGLUE \cite{wang2019superglue} and the first $50$ highest-ranked in SQuAD \cite{rajpurkar2016squad}. LLMs technical reports were collected from MMLU-pro \cite{wang2024mmlu}, from which we gathered all systems that have an aggregated score that is equal or higher than $0.5$, and the first 50 high-ranked from Chatbot Arena \cite{chiang2024chatbot}. Finally, we included all models that have been benchmarked in two existing leaderborards focused on LMs safety: ALERT \cite{tedeschi2024alert} and Secure Learning Lab's LLM safety leaderboard~\footnote{\url{https://huggingface.co/spaces/AI-Secure/llm-trustworthy-leaderboard}}. In this first step we collected $47$ technical reports.

\begin{table*}
\centering
\resizebox{0.6\textwidth}{!}{
        \begin{tabular}{c|p{10cm}}
    \textbf{strategy} & \textbf{models} \\
    \hline
        authoritative source & {\color{red}BERT} \cite{kenton2019bert}, {\color{red}RoBERTa} \cite{liu2019roberta}, {\color{red}XLNet} \cite{yang2019xlnet}, {\color{orange}DeBERTa} \cite{he2021deberta}, {\color{red}ERNIE 3.0} \cite{sun2021ernie}, {\color{green}LinkBERT} \cite{yasunaga2022linkbert}, {\color{orange}Vega v2} \cite{zhong2022toward}\\
        \hline
        document seeding & {\color{orange}GPT-3} \cite{brown2020language}, {\color{green}DSIR} \cite{xie2023data}, {\color{red}Mamba} \cite{gu2023mamba}, {\color{red}MPT-7B} \cite{mosaic2023}, {\color{red}MAmmoTH2} \cite{mickus2024mammoth}\\
        \hline
        quality-based & {\color{orange}GPT-3}\cite{brown2020language}, {\color{red}Gopher} \cite{rae2021scaling}, {\color{red}GLaM} \cite{du2022glam},  {\color{red}Mamba} \cite{gu2023mamba}, {\color{red}MAmmoTH2} \cite{mickus2024mammoth}\\
        \hline
        toxicity classifier & {\color{red}Qwen} \cite{bai2023qwen}, {\color{red}Gemma} \cite{team2024gemma}, {\color{green}OLMO-Dolma} \cite{soldaini-etal-2024-dolma}, {\color{red}Yi-Lighting} \cite{wake2024yi}\\
        \hline
        rule-based & {\color{green}T5-C4} \cite{raffel2020exploring}, {\color{orange}FALCON} \cite{penedo2023refinedweb}, {\color{red}Gemma} \cite{team2024gemma} \\
        \hline
        url blacklists &  {\color{orange}FALCON} \cite{penedo2023refinedweb},\\
        \hline
        human-in-the-loop & {\color{red}LaMDA} \cite{thoppilan2022lamda}, {\color{green}Bloom-ROOTS} \cite{laurenccon2022bigscience}, {\color{green}Phi-3.5} \cite{abdin2024phi}\\
        \hline
        safety policy &  {\color{red}Gemini} \cite{team2023gemini}, {\color{red}EXAONE 3.5} \cite{research2024exaone}, {\color{green}INCITE-RedPajama} \cite{weber2024redpajama}, {\color{red}LLama-3} \cite{dubey2024llama} \\
        
        \hline

        not mentioned & {\color{red}Alpaca} \cite{alpaca}, {\color{red}Claude} \cite{claude23}, {\color{red}GPT-4} \cite{achiam2023gpt}, {\color{red}Zephyr} \cite{tunstall2023zephyr}, {\color{red}DeepSeek-V3} \cite{deepseekai2024deepseekv3technicalreport}, {\color{red}Grok-2} \cite{grok}, {\color{red}Jamda} \cite{lieber2024jamba}, {\color{red}Mixtral} \cite{jiang-etal-2024-jn}, {\color{red}Nova} \cite{nova}, {\color{red}Reka} \cite{team2024reka}
         
    \end{tabular}}
    \caption{A taxonomy of data filtering strategies described in technical reports describing LLMs. Model names are highlighted with different colors depending on the availability of filtering strategies and/or dataset: {\color{green}green} if they are fully available, {\color{orange}orange} if they are partially available, {\color{red}red} if they are not available.}
    \label{tab:taxonomy_data_filtering}
\end{table*}

For each report we checked three types of content: the description of the pretraining dataset, the presentation of data filtering strategies, and the treatment of harmful or biased contents in datasets. In the context of this research, we adopt the term harmful for all the phenomena that might have a negative impact against vulnerable groups to discrimination. In this sense we refer to survey of \cite{blodgett2020language}, who distinguish between \textit{representational} harms, which consist in all the negative representation of groups, and \textit{allocative} harms, which includes all the forms of underrepresentation. 

For each paper we search the following keywords: \textit{data, filter, quality, toxic, bias, hate, stereotype}. This stage led to the retrieval of $13$ additional papers referenced in the sections that cover the description of filtering strategies. Among them, $4$ describe the creation of a pretraining dataset and $1$ a methodology for data sampling. At the end of this process we obtained a pool of $55$ papers for our survey with information about the type of data filtering strategies that have been implemented to remove harmful contents and their availability. 

\subsection{A Taxonomy of Data Filtering Strategies} \label{ss:category}
We defined a taxonomy to categorize filtering strategies in Table \ref{tab:taxonomy_data_filtering}. Given the presence of different LMs belonging to the same family (e.g., LLama, ERNIE), we only listed in the table the latest release of this set of model, unless there are significant changes in data filtering strategies. A change in data filtering is the reason why GPT-4 \cite{achiam2023gpt} is treated separately from GPT-3 \cite{brown2020language} while the previous two instantiations of the model are not mentioned. Whenever the technical report of an LM referred to a dataset created in the context of developing the LLM, such as Olmo \cite{groeneveld-etal-2024-olmo} with the Dolma corpus \cite{soldaini-etal-2024-dolma}, we jointly mentioned them in the table.

\paragraph{Authoritative sources.} This strategy is based on the selection of documents only from authoritative resources validated from the research community. Even if this strategy relies on selection rather than filtering, we mentioned it since it represents a standard approach before the wave of research on the ethical issues related to language modeling \cite{bender2021dangers}. An example of this approach is in BERT \cite{kenton2019bert} that has been trained on a snapshot of Wikipedia and on the Book Corpus \cite{7410368} without implementing any type of quality checks and filters.
\paragraph{Document seeding.} This strategy is based on heuristics for the selection of specific documents from the web. One of the most famous examples of this approach is the one adopted for the creation of the OpenWebText corpus \cite{Gokaslan2019OpenWeb}, which is a collection of all the outbound links from Reddit that have been upvoted at least 3 times. 
\paragraph{Quality-based.} This strategy filters out low-quality documents by comparing them with high-quality documents selected from authoritative sources. For instance, \citet{brown2020language} created a corpus of documents sampled from Wikipedia, OpenWebtext, and the Book Corpus to train a classifier for the removal of documents that are classified as too dissimilar from it. 

\paragraph{Toxicity classifier.} This strategy provides the training of a classifier to filter out all the potentially harmful contents from datasets. The first example of such an approach in our survey is from \citet{gao2020pile} who used profanity-check\footnote{\url{https://pypi.org/project/profanity-check/}} to remove toxic contents from The Pile dataset. Other commonly adopted classifiers are Perspective APIs\footnote{\url{https://perspectiveapi.com/}}, and FastText models trained on corpora for Hate Speech (HS) detection \cite{soldaini-etal-2024-dolma}.
\paragraph{Rule-based.} This strategy exploits lexicons or heuristics to remove unwanted contents. The most studied strategy of this type adopts the Shutterstack Lexicon\footnote{\url{https://bit.ly/3QjwMvz}} \cite{raffel2020exploring} to flag as toxic any sentence containing one of the word in that list. 
\paragraph{Url blacklist.} This strategy removes contents from blacklisted websites. The strategy has been adopted for the creation of the RefinedWeb Dataset \cite{su-etal-2024-refine}, which relied on the semi-automatic creation of a blacklist of urls.
\paragraph{Human-in-the-loop.} This strategy provides the involvement of human evaluators during the creation of datasets. \citet{laurenccon2022bigscience} organized hackatons with NLP communities to create a whitelist of domains to be used to crawl data for pretraining. \citet{thoppilan2022lamda} collected human prompts from crowd-workers aimed at collecting a corpus of Question Answering pairs annotated for harm detection. 
\paragraph{Safety policy.} These strategy relies on the self-assessment of the research team who develops the LM. For instance, the team that developed Gemini declare that they ``perform safety filtering to remove harmful content based on our policies'' \cite{team2023gemini}. \citet{dubey2024llama} detect toxic contents without removing them and use such an information to implement their safety policy in a further step of their language modeling pipeline. 

\subsection{Trends and Limitations in Data Filtering} \label{ss:trends}
In this section we discuss some general trends and limitations that emerge from our overview. 

A first consideration is about the \textbf{the lack of replicability} of filtering strategies presented in technical reports. This trend is shown in Table \ref{tab:taxonomy_data_filtering} where models are highlighted in red if they did not disclose their data filtering pipeline or the snapshot of datasets used for training, in orange if they partially did it, in green if they fully disclose filtering strategies and data. As can be observed, most of the existing LMs are not released with the actual scripts for the replication of implemented filtering methods. Specifically, we recognize a first generation of LMs (e.g.: BERT, RoBERTa, etc.) that have been trained without considering the issue of harmful contents in pretraining datasets. The pivotal work of \citet{bender2021dangers} led to a second generation of models that implemented specific strategies and open-sourced their approaches and results for further investigation. This is the case of the C4 corpus \cite{raffel-etal-2024-simultaneous}, which has been released both in a filtered and unfiltered version. Besides notable exceptions \cite{soldaini-etal-2024-dolma,weber2024redpajama}, the current tendency is to not disclose scripts and results of the data filtering strategies implemented for pretraining. 

A similar trend towards the reduced openness of filtering strategies is observable about the \textbf{documentation debt} \cite{bandy2021addressing} of implemented filtering strategies. If the introduction of documentation templates like Datasheets \cite{gebru2021datasheets} and Model Cards \cite{mitchell2019model} led to a first wave of fully documented self-assessment reports \cite{chowdhery2023palm}, the more recent technical reports formally rely on these documentation templates but substantially provide very generic descriptions about their data filtering approach. For instance, the description provided in the Gemma technical report is limited to the following paragraph: ``We filter the pre-training dataset to reduce the risk of unwanted or unsafe utterances [...]. This includes both heuristics and model-based classifiers to remove harmful or low-quality content'' \cite{team2024gemma}. 

A final consideration is about the \textbf{lack of strategies that are focused on reducing the underrepresentation of groups vulnerable to discrimination} in pretraining datasets. Although it has been shown that filtering strategies for harm reduction have a negative impact on minorities \cite{dodge2021documenting}, few attempts to mitigate this issue have been made so far. A notable exception is the work of \citet{soldaini-etal-2024-dolma}, which provides an assessment of their toxicity classifier over different types of English dialects. However, research works that systematically study the correlation between filtering strategies and increase of minorities underrepresentation have not yet been provided. In Section \ref{sec:experiment} we present a first experiment focused on this issue.

%% file: latex/sections/4.experiments.tex
\section{Measuring the Impact of Filtering Strategies against Vulnerable Groups} \label{sec:experiment}
In this section we benchmark seven filtering strategies on their impact in reducing or increasing the representativeness of vulnerable groups in pretraining datasets. Our experimental setup adopts an intersectional approach \cite{crenshaw2013mapping} as it considers four groups derived from the intersection of people's gender and origin: Western men, Western women, Post-colonial men, Post-colonial women. For this analysis we measure the number of named entities for each demographic group that are removed by the implementation of different filtering strategies, using samples of documents gathered from Common Crawl as a benchmark. In Section \ref{ss:methodology} we describe our experimental setup, in Section \ref{ss:results} we discuss the results of our the analysis

\subsection{Experimental Setting} \label{ss:methodology}
\begin{table*}
    \centering
    \begin{tabular}{c|ccccc}
        &\textbf{strategy} & \textbf{w.m.} & \textbf{p-c.m} & \textbf{w.w.} & \textbf{p-c. w.} \\
        \hline
        &unfiltered & 85276.0 & 15272.0 & 22184.6 & 5292.0 \\
        \hline
       \multirow{2}{*}{rule-based} & shutterstack &  -2.3\%& -1.6\%& \textbf{-4.2\%} & -3\%\\
        &hatebase & -0.4\% & -0.6\%& -0.5\% & \textbf{-0.9\%}\\
        \hline
        \multirow{3}{*}{classifier-based}&perspective & -0.11\% & -0.11\%& \textbf{-0.13\%}& -0.11\% \\
        %perspective_{lang} & -3.2\% & \textbf{-8.3\%}& -3\%& -6.3\% \\

        &fasttext & -0.3\%&-0.3\% & \textbf{-0.9\%}& -0.7\%\\
       & profanity-check & -0.21\%&-0.22\% & \textbf{-0.89\%}& -0.52\%\\
        \hline
       \multirow{2}{*}{quality-based} & quality\_wiki & -15.1\% & \textbf{-17.2\%} & -12.7\% & -11.2\% \\
        &quality\_webtext & \textbf{-44.6\%} & -42.6\% & -33.1\% & -33.4\% \  
    \end{tabular}
    \caption{The average number of named entities in the 5 samples gathered from Common Crawl and the percentage of sentences removed by applying $7$ filtering strategies. Named entities are broken down by groups resulting from the intersection of gender and origin: Western men (w.m.), Post-colonial men (p-c.m.), Western women (w.-w.), and Post-colonial women (p-c.-w.)}
    \label{tab:experimental_results}
\end{table*}

\paragraph{Knowledge base creation.} The first step of our experimental setting was the creation of a knowledge base that enables the categorization of named entities on the basis of their gender and origin. In order to do so we developed the \textbf{People Dataset}, a corpus of $10.8$ million of entities of the type person with information about their country of birth, citizenship, gender, ethnic minority, and occupation. The dataset is derived from Wikidata \cite{erxleben2014introducing}, a collaborative Knowledge Graph (KG) maintained by the Wikimedia ecosystem, and processed on the basis of previous literature on the topic \cite{stranisci2023wikibio}: we inferred people's countries of birth from Wikidata property `place of birth' (P19) properties, mapped all the properties of the type `ethnic group' (P172) with a curated list of Post-colonial minorities in Western countries (e.g., African-Americans). Finally, we integrated the knowledge base with additional information from CaLiGraph \cite{heist2019uncovering}, a KG derived from Wikipedia categories.

\paragraph{Sampling from Common Crawl.} We gathered $5$ samples of documents from the Common Crawl (CC) snapshot released in August 2024\footnote{\url{https://data.commoncrawl.org/crawl-data/CC-MAIN-2024-33/index.html}}. Each sample is composed of $20$ Web ARChive (WARC) files that have been processed according to the following criteria: \textit{i.} we kept only documents classified as written in English, implementing the langdetect Python library\footnote{\url{https://pypi.org/project/langdetect/}}; \textit{ii.} we removed all the documents with an average number of sentences below $5$ and an average number of words \textit{per} sentence below $5$. The average number of documents in each sample is $102,820$ (std: $443$).

\paragraph{Entity Linking.} The next step of our approach was the linking of named entities with the People Dataset. In order to adopt a computationally-efficient Entity Linking (EL) approach to implement on a large set of data, we designed an EL pipeline that combines the adoption of neural methods for the detection of named entities and existing heuristics to link them to the dataset \cite{stranisci2023world}. The pipeline was organized in four steps. 

\begin{enumerate}
    \item We used the largest SpaCy model\footnote{\url{https://spacy.io/, en\_core\_web\_sm}} to detect all the named entities of the type PERSON in CC samples.
    \item We queried Wikipedia APIs\footnote{\url{https://www.mediawiki.org/wiki/API:Search}} with all the retrieved named entities to obtain titles of Wikipedia pages and filtered out all the named entities that were too dissimilar from Wikipedia titles, through the adoption of a set of heuristics already validated by \citet{manghi2023miriam}.
    \item We retrieved all the Wikidata IDs corresponding to the Wikipedia 
    titles.
    \item We kept only entities with a Wikidata ID that is present in our People Dataset.
\end{enumerate}
  As a result (Table \ref{tab:experimental_results}), we identified an average of $128,025$ entities from each sample: $85,276$ Western men, $15,272$ Post-colonial men, $22,185$ Western women, $5,292$ Post-colonial women. In order to assess whether there is high variation between samples we conducted an ANOVA test \cite{st1989analysis} on all the possible pairs of entities distributions broken-down by group. In all cases we obtained a $p-value$ of $0.99$, showing that there is no evidence of variation between different CC samples. 

\paragraph{Implementation of Filtering Strategies.} Having created our baseline, we implemented seven filtering strategies belonging to three different approaches outlined in the taxonomy \ref{tab:taxonomy_data_filtering}, which can be replicated in an experimental setting: rule-based, toxicity classifier, and document similarity. These strategies were applied to all the sentences or documents that mentioned at least one of the retrieved named entities.

The implementation of rule-based strategies relies on two different lexicons for the identification of toxic language: the Shutterstack lexicon and a version of HateBase that has been refined by \cite{hateoffensive} from their Hate Speech corpus\footnote{\url{https://github.com/t-davidson/hate-speech-and-offensive-language/tree/master/lexicons}
}. We flagged as harmful any sentence that contains at least one term included in each lexicon. 

For the comparison of model-based strategies we adopted three toxicity classifiers: Perspective API, which has been used during the training of models like Gemini \cite{team2023gemini} and Gopher \cite{rae2021scaling}, profanity-check\footnote{\url{https://pypi.org/project/alt-profanity-check/}}, which has been used to filter the Pile Corpus \cite{gao2020pile}, a FastText classifier trained on the Jigsaw dataset \cite{Jigsaw} that has been used to polish Dolma\footnote{\url{https://github.com/allenai/dolma/blob/main/python/dolma/\\taggers/jigsaw.py}} \cite{soldaini-etal-2024-dolma}. We considered all sentences classified containing Hate Speech by the FastText classifier as harmful. Since Perspective APIs and profanity-check output a probability distribution, we filtered out all sentences classified as toxic with a probability of $0.8$ or more, a threshold in line with previous research \cite{xu2021detoxifying}. 

For the document similarity strategy we replicated the methodology proposed by \cite{brown2020language} and \cite{du2022glam}. We trained a hash based-linear classifier with a vector-size of $1,000$ on a corpus of high-quality documents to filter out documents that do not reach a given quality threshold. Classifiers were trained on a total of $100,000$ documents: $50,000$ considered as high-quality (positive class), $50,000$ considered as low-quality, which were randomly gathered from CC WARC files that were not used during our sampling step. We chose two sources of high-quality documents: the first based on a random sample of Wikipedia documents, the second on a sample from the OpenWebText corpus \cite{Gokaslan2019OpenWeb}.\footnote{The People Dataset and all the implemented filtered strategies will be released under MIT license after the anonymity period.}

\subsection{Analysis of Results} \label{ss:results}
Table \ref{tab:experimental_results} shows the results of our experiments. The first row of the table reports the average number of entity mentions in samples. In the other rows the impact of a data filtering strategy on named entities is reported in terms of the percentage of removed mentions from the sample. Filtering strategies are grouped by the taxonomy they belong to (Section \ref{ss:category}) and results are broken-down by sociodemographic group: Western men (w.m.), Western women (w.w.), Post-colonial men (p-c.m.), and Post-colonial women (p-c.w.). 

\paragraph{How much is filtered out?} From a general overview of results it is possible to observe that strategies have a widely varying impact on our samples. The filtering strategy based on the Hatebase lexicon \cite{davidson-etal-2024-self} and all classifier-based approaches have a marginal effect on documents, filtering out less that $1\%$ of sentences. The Shutterstack lexicon has a greater impact, spanning between $1\%$ and $4.2\%$ of removed mentions. Document-similarity approaches lead to a more aggressive filtering that ranges between $11.2\%$ and $44.6\%$ of removed contents.

\paragraph{What is filtered out?} The magnitude of filtering strategies' impact is not fully explicative of their behavior without considering which texts they flag for the removal. For each category of filtering strategy we performed an analysis aimed at understanding which pattern they follow for removing contents. To investigate rule-based filtering strategies we obtained the distribution of the lexical items that have been found in all samples and identified the five most frequent ones for each lexicon (Table \ref{tab:top-k matching words}). The comparison shows two different filtering patterns between the two strategies. The most-frequent words from Shutterstack lexicon focus on pornography (e.g., `sex'); HateBase terms on racism (e.g., `slaves', `blacks') and misogyny (e.g., `dykes'). Choosing one of the two lexicons does not have only an impact in the number of flagged sentences but also on the specific subset of potentially harmful contents that are removed. 

\begin{table}
    \centering
    \begin{tabular}{cp{5cm}}
        \textbf{lexicon} & \textbf{top-5}\\
        \hline
        shutterstack & dick, sex, porn, ass, nude\\
        hatebase & slave, married to, blacks, dykes, of white\\
    \end{tabular}
    \caption{Top-5 matching lexical items in CC samples}
    \label{tab:top-k matching words}
\end{table}

\begin{figure}
    \centering
    \includegraphics[width=0.9\columnwidth]{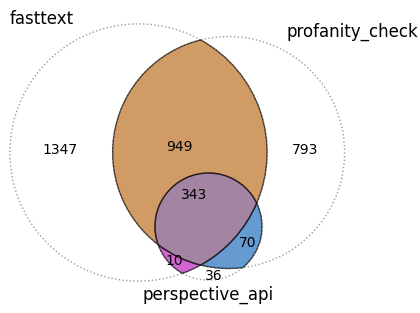}
    \caption{Intersection of contents removed through classifier-based strategies}
    \label{fig:venn}
\end{figure}

The comparison of classifier-based strategies focuses on the number of flagged sentences in which their classification overlaps and those where it does not. 
In Figure \ref{fig:venn}, we can identify two patterns. The total number of sentences classified as harmful by Perspective API is almost a subset of ones classified by Profanity Check with an overlap of $90\%$. The overlap of Perspective API with the FastText classifier is lower but still significant ($77\%$). This pattern can be explained by the small number of sentences flagged by Perspective API ($459$) compared to Profanity Check ($2,155$) and FastText ($2,649$). The second pattern is the strong difference in the sentences identified by Profanity Check and FastText: only $35.8\%$ of sentences classified as harmful by FastText are also classified as such by Profanity Check. This shows that, similarly to the lexicon-based strategies, choosing one of the existing classifiers for harm detection may imply targeting different types of harmfulness.  

Quality-based filtering strategies do not directly aim at harm detection but their training over allegedly high-quality documents is supposed to have an impact on the removal of toxic contents from raw documents. We check their effectiveness by counting the percentage of mentions kept by these strategies despite having been flagged as harmful by rule-based and classifier-based approaches. Results in Table \ref{tab:kept_harms} show that a high percentage of sentences classified as harmful are still present after the quality-based filtering. Adopting the classifier trained on Wikipedia documents leads to the removal of $15\%$ of sentences while keeping $93.5\%$ of harmful sentences. The classifier based on OpenWebText shows a similar proportion: it removes $45\%$ of sentences but $69.3\%$ of sentences classified as harmful by other strategies are still present after the filtering. This comparison reveals that quality is not a proxy of safety, since the most part of toxic contents is not removed through the adoption of quality-based strategies.  

\begin{table}
    \centering
    \begin{tabular}{ccc}
        \textbf{strategy} & \textbf{n. sents (\%)} & toxic sents (\%)\\
        \hline
        wikipedia & 85\% & $90.5\%$\\
        webtext & 55\% & $69.3\%$\\
    \end{tabular}
    \caption{The percentage of harmful contents that remain after the application of quality-based strategies}
    \label{tab:kept_harms}
\end{table}

\paragraph{Who gets filtered out?} A third line of analysis studies what entities are most impacted by filtering strategies. As can be seen in Table \ref{tab:experimental_results}, the adoption of different strategies not only differs in its magnitude but systematically penalizes certain categories of people. Women are always the most impacted target in rule-based and classifier-based strategies, and in $4$ cases out of $5$, named entities that suffer the highest content removal are Western women. Conversely, strategies based on document similarity have diametrically opposed effect, since they impact the most on men. However, as shown in Table \ref{tab:kept_harms}, filtering based on document quality is not a reliable method of identifying toxic contents. Therefore, it is not possible to assume that these strategies remove entity mentions that appear in harmful contents and their major impact on men cannot be interpreted as a side-effect of harm detection.

In order to deepen our analysis of the impact of data filtering strategies against groups, we leveraged our People Dataset to identify which are the occupations of named entities that are removed through data filtering strategies. For each entity mention occurring in a sentence that has been flagged as harmful by a rule-based or a classifier-based strategy we retrieved their occupation from our dataset and computed the occupations that occurred the most for each analyzed group. Since we observed that quality-based filtering strategies are not effective in the detection of toxic content, we did not consider them in this analysis. 

Table \ref{tab:professions} shows the $5$ most occurring occupations in datasets and the $5$ that have been mostly flagged for removal by filtering strategies. The analysis confirms the presence of patterns of discrimination along the gender axis. The distribution of men's occupations that have been filtered is coherent with the original distribution of mentions; for women this is not the same. The most removed occupation of Western women is `pornographic actor', which is not among the most frequent occupations of Western women. Similarly, the profession `model' is one of the most removed from post-colonial women occupations despite not being frequent in the original distribution. This means that the removal of specific forms of harm, which has not necessarily a negative effect, increases the underrepresentation of very specific categories of people, suggesting the need of adopting finer-grained analysis of the impact of filtering strategies on people. 

\begin{table}
    \centering
    \resizebox{0.9\columnwidth}{!}{
    \begin{tabular}{cp{3cm}p{3cm}}
         \textbf{group} & \textbf{top-k occupation}& \textbf{top-k filtered}\\
         \hline
         w.m. & writer, politician, actor, film actor, television actor &actor, writer, film actor, politician, television actor \\
         p-c. m. & politician, actor, writer, television actor, singer &  politician, actor, singer, writer, film actor \\
        w. w. & actor, film actor, singer, television actor, writer & pornographic actor, actor, film actor, film director, television actor\\
        p-c. w. & actor, politician, writer, film actor, singer &  actor, film actor, singer, model, writer \\
    \end{tabular}}
    \caption{top-5 removal by profession}
    \label{tab:professions}
\end{table}

%% file: latex/sections/5.discussion.tex
\section{Discussion}
Our survey of data filtering strategies shows that the implementation of effective approaches to detect and remove harmful contents from pretraining datasets is still an open issue with important social implication. Choosing a strategy means addressing a specific subcategory of harmfulness and this has an impact against specific groups of people (RQ 2). Despite this variety, a general pattern emerges from the implementation of all rule-based and classifier-based strategies:\textbf{ the systematic increase of underrepresentation of women in pretraining datasets}. Mentions of women are always removed at a highest rate than men. If at a first look quality-based strategies seems to have a highest impact against men, we discovered that they do not necessarily remove harmful contents. This implies that relying on allegedly neutral sources of knowledge like Wikipedia and the OpenWebText corpus does not prevent from keeping harmful contents against certain groups of people. 

In contrast with this evidence about the complex nature of data filtering for harm reduction, the general tendency that emerges from our survey (RQ 1) is disengagement with this issue. After a wave of efforts in this field connected with the critical work of \cite{bender2021dangers}, \textbf{the interest in implementing robust pretraining data filtering dramatically reduced} in favor of post-training measures for harm reduction. In most cases, the actual implementation of strategies is close-sourced, hindering the replication of procedures that are adopted for processing datasets. The presence of notable exceptions such as Dolma \cite{soldaini-etal-2024-dolma} and the RefinedWeb Dataset \cite{penedo2023refinedweb} represents a significant counter-tendency. However, there is still a lack of filtering strategies that are effective in balancing the need to remove harmful contents while preserving the representativeness of vulnerable groups in pretraining datasets.

%% file: latex/sections/6.conclusion.tex
\section{Conclusion}
In this paper we presented a first systematic analysis of data filtering strategies for harm reduction in datasets. The overview of $55$ technical reports describing LLMs enabled us to draw a taxonomy of filtering strategies and to identify the open issues that prevent the implementation of effective, reliable, and replicable methods for the reduction of harmful contents. Additionally, we experimentally evaluated seven existing strategies on CC samples of documents in order to assess their impact against groups characterized by the intersection of gender and origin. The evaluation showed a systematic negative effect of strategies against women and a high variety in the types of harmful contents that specific strategies filter out.

Future work will focus on providing a more effective pipeline for the assessment of data filtering strategies with three aims in mind: \textit{i.} improving the coverage of our pipeline by including additional information in the People Dataset. We plan to include information from additional knowledge bases and add information about vulnerable groups that goes beyond named entities (e.g., mentions of communities, ethnic groups, and demonyms) ; \textit{ii.} adopting a participatory approach to the evaluation of filtering strategies that engages associations and communities of people who are active against discrimination. This approach will enable to explore the data filtering process from the perspectives of people who might actually harmed by the implementation of strategies; \textit{iii} we will systematically explore the correlation between the adoption of specific data filtering strategies and linguistic behaviors emerging from models trained on filtered datasets.

%% file: acl_latex.bbl
\begin{thebibliography}{74}
\providecommand{\natexlab}[1]{#1}

\bibitem[{Abdin et~al.(2024)Abdin, Aneja, Awadalla, Awadallah, Awan, Bach, Bahree, Bakhtiari, Bao, Behl et~al.}]{abdin2024phi}
Marah Abdin, Jyoti Aneja, Hany Awadalla, Ahmed Awadallah, Ammar~Ahmad Awan, Nguyen Bach, Amit Bahree, Arash Bakhtiari, Jianmin Bao, Harkirat Behl, et~al. 2024.
\newblock Phi-3 technical report: A highly capable language model locally on your phone.
\newblock \emph{arXiv preprint arXiv:2404.14219}.

\bibitem[{Achiam et~al.(2023)Achiam, Adler, Agarwal, Ahmad, Akkaya, Aleman, Almeida, Altenschmidt, Altman, Anadkat et~al.}]{achiam2023gpt}
Josh Achiam, Steven Adler, Sandhini Agarwal, Lama Ahmad, Ilge Akkaya, Florencia~Leoni Aleman, Diogo Almeida, Janko Altenschmidt, Sam Altman, Shyamal Anadkat, et~al. 2023.
\newblock Gpt-4 technical report.
\newblock \emph{arXiv preprint arXiv:2303.08774}.

\bibitem[{Albalak et~al.(2024)Albalak, Elazar, Xie, Longpre, Lambert, Wang, Muennighoff, Hou, Pan, Jeong et~al.}]{albalak2024survey}
Alon Albalak, Yanai Elazar, Sang~Michael Xie, Shayne Longpre, Nathan Lambert, Xinyi Wang, Niklas Muennighoff, Bairu Hou, Liangming Pan, Haewon Jeong, et~al. 2024.
\newblock A survey on data selection for language models.
\newblock \emph{arXiv preprint arXiv:2402.16827}.

\bibitem[{Antropic(2023)}]{claude23}
Antropic. 2023.
\newblock Model card and evaluations for claude models.
\newblock \url{https://www-cdn.anthropic.com/files/4zrzovbb/website/bd2a28d2535bfb0494cc8e2a3bf135d2e7523226.pdf }.

\bibitem[{Bai et~al.(2023)Bai, Bai, Chu, Cui, Dang, Deng, Fan, Ge, Han, Huang et~al.}]{bai2023qwen}
Jinze Bai, Shuai Bai, Yunfei Chu, Zeyu Cui, Kai Dang, Xiaodong Deng, Yang Fan, Wenbin Ge, Yu~Han, Fei Huang, et~al. 2023.
\newblock Qwen technical report.
\newblock \emph{arXiv preprint arXiv:2309.16609}.

\bibitem[{Bandy and Vincent(2021)}]{bandy2021addressing}
Jack Bandy and Nicholas Vincent. 2021.
\newblock Addressing" documentation debt" in machine learning: A retrospective datasheet for bookcorpus.
\newblock In \emph{Thirty-fifth Conference on Neural Information Processing Systems Datasets and Benchmarks Track (Round 1)}.

\bibitem[{Bender et~al.(2021)Bender, Gebru, McMillan-Major, and Shmitchell}]{bender2021dangers}
Emily~M Bender, Timnit Gebru, Angelina McMillan-Major, and Shmargaret Shmitchell. 2021.
\newblock On the dangers of stochastic parrots: Can language models be too big?
\newblock In \emph{Proceedings of the 2021 ACM conference on fairness, accountability, and transparency}, pages 610--623.

\bibitem[{Blodgett et~al.(2020)Blodgett, Barocas, Daum{\'e}~III, and Wallach}]{blodgett2020language}
Su~Lin Blodgett, Solon Barocas, Hal Daum{\'e}~III, and Hanna Wallach. 2020.
\newblock Language (technology) is power: A critical survey of “bias” in nlp.
\newblock In \emph{Proceedings of the 58th Annual Meeting of the Association for Computational Linguistics}, pages 5454--5476.

\bibitem[{Brown et~al.(2020)Brown, Mann, Ryder, Subbiah, Kaplan, Dhariwal, Neelakantan, Shyam, Sastry, Askell et~al.}]{brown2020language}
Tom Brown, Benjamin Mann, Nick Ryder, Melanie Subbiah, Jared~D Kaplan, Prafulla Dhariwal, Arvind Neelakantan, Pranav Shyam, Girish Sastry, Amanda Askell, et~al. 2020.
\newblock Language models are few-shot learners.
\newblock \emph{Advances in neural information processing systems}, 33:1877--1901.

\bibitem[{Chiang et~al.(2024)Chiang, Zheng, Sheng, Angelopoulos, Li, Li, Zhang, Zhu, Jordan, Gonzalez et~al.}]{chiang2024chatbot}
Wei-Lin Chiang, Lianmin Zheng, Ying Sheng, Anastasios~Nikolas Angelopoulos, Tianle Li, Dacheng Li, Hao Zhang, Banghua Zhu, Michael Jordan, Joseph~E Gonzalez, et~al. 2024.
\newblock Chatbot arena: An open platform for evaluating llms by human preference.
\newblock \emph{arXiv preprint arXiv:2403.04132}.

\bibitem[{Chowdhery et~al.(2023)Chowdhery, Narang, Devlin, Bosma, Mishra, Roberts, Barham, Chung, Sutton, Gehrmann et~al.}]{chowdhery2023palm}
Aakanksha Chowdhery, Sharan Narang, Jacob Devlin, Maarten Bosma, Gaurav Mishra, Adam Roberts, Paul Barham, Hyung~Won Chung, Charles Sutton, Sebastian Gehrmann, et~al. 2023.
\newblock Palm: Scaling language modeling with pathways.
\newblock \emph{Journal of Machine Learning Research}, 24(240):1--113.

\bibitem[{Crenshaw(2013)}]{crenshaw2013mapping}
Kimberl{\'e}~Williams Crenshaw. 2013.
\newblock Mapping the margins: Intersectionality, identity politics, and violence against women of color.
\newblock In \emph{The public nature of private violence}, pages 93--118. Routledge.

\bibitem[{Davidson et~al.(2017)Davidson, Warmsley, Macy, and Weber}]{hateoffensive}
Thomas Davidson, Dana Warmsley, Michael Macy, and Ingmar Weber. 2017.
\newblock Automated hate speech detection and the problem of offensive language.
\newblock In \emph{Proceedings of the 11th International AAAI Conference on Web and Social Media}, ICWSM '17, pages 512--515.

\bibitem[{Davidson et~al.(2024)Davidson, Surkov, Veselovsky, Russo, West, and Gulcehre}]{davidson-etal-2024-self}
Tim~R. Davidson, Viacheslav Surkov, Veniamin Veselovsky, Giuseppe Russo, Robert West, and Caglar Gulcehre. 2024.
\newblock \href {https://doi.org/10.18653/v1/2024.findings-emnlp.703} {Self-recognition in language models}.
\newblock In \emph{Findings of the Association for Computational Linguistics: EMNLP 2024}, pages 12032--12059, Miami, Florida, USA. Association for Computational Linguistics.

\bibitem[{DeepSeek-AI et~al.(2024)DeepSeek-AI, Liu, Feng, Xue, Wang, Wu, Lu, Zhao, Deng, Zhang, Ruan, Dai, Guo, Yang, Chen, Ji, Li, Lin, Dai, Luo, Hao, Chen, Li, Zhang, Bao, Xu, Wang, Zhang, Ding, Xin, Gao, Li, Qu, Cai, Liang, Guo, Ni, Li, Wang, Chen, Chen, Yuan, Qiu, Li, Song, Dong, Hu, Gao, Guan, Huang, Yu, Wang, Zhang, Xu, Xia, Zhao, Wang, Zhang, Li, Wang, Zhang, Zhang, Tang, Li, Tian, Huang, Wang, Zhang, Wang, Zhu, Chen, Du, Chen, Jin, Ge, Zhang, Pan, Wang, Xu, Zhang, Chen, Li, Lu, Zhou, Chen, Wu, Ye, Ye, Ma, Wang, Zhou, Yu, Zhou, Pan, Wang, Yun, Pei, Sun, Xiao, Zeng, Zhao, An, Liu, Liang, Gao, Yu, Zhang, Li, Jin, Wang, Bi, Liu, Wang, Shen, Chen, Zhang, Chen, Nie, Sun, Wang, Cheng, Liu, Xie, Liu, Yu, Song, Shan, Zhou, Yang, Li, Su, Lin, Li, Wang, Wei, Zhu, Zhang, Xu, Xu, Huang, Li, Zhao, Sun, Li, Wang, Yu, Zheng, Zhang, Shi, Xiong, He, Tang, Piao, Wang, Tan, Ma, Liu, Guo, Wu, Ou, Zhu, Wang, Gong, Zou, He, Zha, Xiong, Ma, Yan, Luo, You, Liu, Zhou, Wu, Ren, Ren, Sha, Fu, Xu, Huang, Zhang, Xie, Zhang, Hao,
  Gou, Ma, Yan, Shao, Xu, Wu, Zhang, Li, Gu, Zhu, Liu, Li, Xie, Song, Gao, and Pan}]{deepseekai2024deepseekv3technicalreport}
DeepSeek-AI, Aixin Liu, Bei Feng, Bing Xue, Bingxuan Wang, Bochao Wu, Chengda Lu, Chenggang Zhao, Chengqi Deng, Chenyu Zhang, Chong Ruan, Damai Dai, Daya Guo, Dejian Yang, Deli Chen, Dongjie Ji, Erhang Li, Fangyun Lin, Fucong Dai, Fuli Luo, Guangbo Hao, Guanting Chen, Guowei Li, H.~Zhang, Han Bao, Hanwei Xu, Haocheng Wang, Haowei Zhang, Honghui Ding, Huajian Xin, Huazuo Gao, Hui Li, Hui Qu, J.~L. Cai, Jian Liang, Jianzhong Guo, Jiaqi Ni, Jiashi Li, Jiawei Wang, Jin Chen, Jingchang Chen, Jingyang Yuan, Junjie Qiu, Junlong Li, Junxiao Song, Kai Dong, Kai Hu, Kaige Gao, Kang Guan, Kexin Huang, Kuai Yu, Lean Wang, Lecong Zhang, Lei Xu, Leyi Xia, Liang Zhao, Litong Wang, Liyue Zhang, Meng Li, Miaojun Wang, Mingchuan Zhang, Minghua Zhang, Minghui Tang, Mingming Li, Ning Tian, Panpan Huang, Peiyi Wang, Peng Zhang, Qiancheng Wang, Qihao Zhu, Qinyu Chen, Qiushi Du, R.~J. Chen, R.~L. Jin, Ruiqi Ge, Ruisong Zhang, Ruizhe Pan, Runji Wang, Runxin Xu, Ruoyu Zhang, Ruyi Chen, S.~S. Li, Shanghao Lu, Shangyan Zhou, Shanhuang
  Chen, Shaoqing Wu, Shengfeng Ye, Shengfeng Ye, Shirong Ma, Shiyu Wang, Shuang Zhou, Shuiping Yu, Shunfeng Zhou, Shuting Pan, T.~Wang, Tao Yun, Tian Pei, Tianyu Sun, W.~L. Xiao, Wangding Zeng, Wanjia Zhao, Wei An, Wen Liu, Wenfeng Liang, Wenjun Gao, Wenqin Yu, Wentao Zhang, X.~Q. Li, Xiangyue Jin, Xianzu Wang, Xiao Bi, Xiaodong Liu, Xiaohan Wang, Xiaojin Shen, Xiaokang Chen, Xiaokang Zhang, Xiaosha Chen, Xiaotao Nie, Xiaowen Sun, Xiaoxiang Wang, Xin Cheng, Xin Liu, Xin Xie, Xingchao Liu, Xingkai Yu, Xinnan Song, Xinxia Shan, Xinyi Zhou, Xinyu Yang, Xinyuan Li, Xuecheng Su, Xuheng Lin, Y.~K. Li, Y.~Q. Wang, Y.~X. Wei, Y.~X. Zhu, Yang Zhang, Yanhong Xu, Yanhong Xu, Yanping Huang, Yao Li, Yao Zhao, Yaofeng Sun, Yaohui Li, Yaohui Wang, Yi~Yu, Yi~Zheng, Yichao Zhang, Yifan Shi, Yiliang Xiong, Ying He, Ying Tang, Yishi Piao, Yisong Wang, Yixuan Tan, Yiyang Ma, Yiyuan Liu, Yongqiang Guo, Yu~Wu, Yuan Ou, Yuchen Zhu, Yuduan Wang, Yue Gong, Yuheng Zou, Yujia He, Yukun Zha, Yunfan Xiong, Yunxian Ma, Yuting Yan, Yuxiang
  Luo, Yuxiang You, Yuxuan Liu, Yuyang Zhou, Z.~F. Wu, Z.~Z. Ren, Zehui Ren, Zhangli Sha, Zhe Fu, Zhean Xu, Zhen Huang, Zhen Zhang, Zhenda Xie, Zhengyan Zhang, Zhewen Hao, Zhibin Gou, Zhicheng Ma, Zhigang Yan, Zhihong Shao, Zhipeng Xu, Zhiyu Wu, Zhongyu Zhang, Zhuoshu Li, Zihui Gu, Zijia Zhu, Zijun Liu, Zilin Li, Ziwei Xie, Ziyang Song, Ziyi Gao, and Zizheng Pan. 2024.
\newblock \href {https://arxiv.org/abs/2412.19437} {Deepseek-v3 technical report}.
\newblock \emph{Preprint}, arXiv:2412.19437.

\bibitem[{Dodge et~al.(2021)Dodge, Sap, Marasovi{\'c}, Agnew, Ilharco, Groeneveld, Mitchell, and Gardner}]{dodge2021documenting}
Jesse Dodge, Maarten Sap, Ana Marasovi{\'c}, William Agnew, Gabriel Ilharco, Dirk Groeneveld, Margaret Mitchell, and Matt Gardner. 2021.
\newblock Documenting large webtext corpora: A case study on the colossal clean crawled corpus.
\newblock In \emph{Proceedings of the 2021 Conference on Empirical Methods in Natural Language Processing}, pages 1286--1305.

\bibitem[{Du et~al.(2022)Du, Huang, Dai, Tong, Lepikhin, Xu, Krikun, Zhou, Yu, Firat et~al.}]{du2022glam}
Nan Du, Yanping Huang, Andrew~M Dai, Simon Tong, Dmitry Lepikhin, Yuanzhong Xu, Maxim Krikun, Yanqi Zhou, Adams~Wei Yu, Orhan Firat, et~al. 2022.
\newblock Glam: Efficient scaling of language models with mixture-of-experts.
\newblock In \emph{International Conference on Machine Learning}, pages 5547--5569. PMLR.

\bibitem[{Dubey et~al.(2024)Dubey, Jauhri, Pandey, Kadian, Al-Dahle, Letman, Mathur, Schelten, Yang, Fan et~al.}]{dubey2024llama}
Abhimanyu Dubey, Abhinav Jauhri, Abhinav Pandey, Abhishek Kadian, Ahmad Al-Dahle, Aiesha Letman, Akhil Mathur, Alan Schelten, Amy Yang, Angela Fan, et~al. 2024.
\newblock The llama 3 herd of models.
\newblock \emph{arXiv preprint arXiv:2407.21783}.

\bibitem[{Edwards(2021)}]{edwards2021eu}
Lilian Edwards. 2021.
\newblock The eu ai act: a summary of its significance and scope.
\newblock \emph{Artificial Intelligence (the EU AI Act)}, 1.

\bibitem[{Erxleben et~al.(2014)Erxleben, G{\"u}nther, Kr{\"o}tzsch, Mendez, and Vrande{\v{c}}i{\'c}}]{erxleben2014introducing}
Fredo Erxleben, Michael G{\"u}nther, Markus Kr{\"o}tzsch, Julian Mendez, and Denny Vrande{\v{c}}i{\'c}. 2014.
\newblock Introducing wikidata to the linked data web.
\newblock In \emph{The Semantic Web--ISWC 2014: 13th International Semantic Web Conference, Riva del Garda, Italy, October 19-23, 2014. Proceedings, Part I 13}, pages 50--65. Springer.

\bibitem[{Gao et~al.(2020)Gao, Biderman, Black, Golding, Hoppe, Foster, Phang, He, Thite, Nabeshima et~al.}]{gao2020pile}
Leo Gao, Stella Biderman, Sid Black, Laurence Golding, Travis Hoppe, Charles Foster, Jason Phang, Horace He, Anish Thite, Noa Nabeshima, et~al. 2020.
\newblock The pile: An 800gb dataset of diverse text for language modeling.
\newblock \emph{arXiv preprint arXiv:2101.00027}.

\bibitem[{Gebru et~al.(2021)Gebru, Morgenstern, Vecchione, Vaughan, Wallach, Iii, and Crawford}]{gebru2021datasheets}
Timnit Gebru, Jamie Morgenstern, Briana Vecchione, Jennifer~Wortman Vaughan, Hanna Wallach, Hal~Daum{\'e} Iii, and Kate Crawford. 2021.
\newblock Datasheets for datasets.
\newblock \emph{Communications of the ACM}, 64(12):86--92.

\bibitem[{Gehman et~al.(2020)Gehman, Gururangan, Sap, Choi, and Smith}]{gehman2020realtoxicityprompts}
Samuel Gehman, Suchin Gururangan, Maarten Sap, Yejin Choi, and Noah~A Smith. 2020.
\newblock Realtoxicityprompts: Evaluating neural toxic degeneration in language models.
\newblock In \emph{Findings of the Association for Computational Linguistics: EMNLP 2020}, pages 3356--3369.

\bibitem[{Gemini et~al.(2023)Gemini, Anil, Borgeaud, Alayrac, Yu, Soricut, Schalkwyk, Dai, Hauth, Millican et~al.}]{team2023gemini}
Team Gemini, Rohan Anil, Sebastian Borgeaud, Jean-Baptiste Alayrac, Jiahui Yu, Radu Soricut, Johan Schalkwyk, Andrew~M Dai, Anja Hauth, Katie Millican, et~al. 2023.
\newblock Gemini: a family of highly capable multimodal models.
\newblock \emph{arXiv preprint arXiv:2312.11805}.

\bibitem[{Gokaslan et~al.(2019)Gokaslan, Cohen, Pavlick, and Tellex}]{Gokaslan2019OpenWeb}
Aaron Gokaslan, Vanya Cohen, Ellie Pavlick, and Stefanie Tellex. 2019.
\newblock Openwebtext corpus.
\newblock \url{http://Skylion007.github.io/OpenWebTextCorpus}.

\bibitem[{Groeneveld et~al.(2024)Groeneveld, Beltagy, Walsh, Bhagia, Kinney, Tafjord, Jha, Ivison, Magnusson, Wang, Arora, Atkinson, Authur, Chandu, Cohan, Dumas, Elazar, Gu, Hessel, Khot, Merrill, Morrison, Muennighoff, Naik, Nam, Peters, Pyatkin, Ravichander, Schwenk, Shah, Smith, Strubell, Subramani, Wortsman, Dasigi, Lambert, Richardson, Zettlemoyer, Dodge, Lo, Soldaini, Smith, and Hajishirzi}]{groeneveld-etal-2024-olmo}
Dirk Groeneveld, Iz~Beltagy, Evan Walsh, Akshita Bhagia, Rodney Kinney, Oyvind Tafjord, Ananya Jha, Hamish Ivison, Ian Magnusson, Yizhong Wang, Shane Arora, David Atkinson, Russell Authur, Khyathi Chandu, Arman Cohan, Jennifer Dumas, Yanai Elazar, Yuling Gu, Jack Hessel, Tushar Khot, William Merrill, Jacob Morrison, Niklas Muennighoff, Aakanksha Naik, Crystal Nam, Matthew Peters, Valentina Pyatkin, Abhilasha Ravichander, Dustin Schwenk, Saurabh Shah, William Smith, Emma Strubell, Nishant Subramani, Mitchell Wortsman, Pradeep Dasigi, Nathan Lambert, Kyle Richardson, Luke Zettlemoyer, Jesse Dodge, Kyle Lo, Luca Soldaini, Noah Smith, and Hannaneh Hajishirzi. 2024.
\newblock \href {https://doi.org/10.18653/v1/2024.acl-long.841} {{OLM}o: Accelerating the science of language models}.
\newblock In \emph{Proceedings of the 62nd Annual Meeting of the Association for Computational Linguistics (Volume 1: Long Papers)}, pages 15789--15809, Bangkok, Thailand. Association for Computational Linguistics.

\bibitem[{Gu and Dao(2023)}]{gu2023mamba}
Albert Gu and Tri Dao. 2023.
\newblock Mamba: Linear-time sequence modeling with selective state spaces.
\newblock \emph{arXiv preprint arXiv:2312.00752}.

\bibitem[{He et~al.(2021)He, Liu, Gao, and Chen}]{he2021deberta}
Pengcheng He, Xiaodong Liu, Jianfeng Gao, and Weizhu Chen. 2021.
\newblock \href {https://openreview.net/forum?id=XPZIaotutsD} {Deberta: Decoding-enhanced bert with disentangled attention}.
\newblock In \emph{International Conference on Learning Representations}.

\bibitem[{Heist and Paulheim(2019)}]{heist2019uncovering}
Nicolas Heist and Heiko Paulheim. 2019.
\newblock Uncovering the semantics of wikipedia categories.
\newblock In \emph{The Semantic Web--ISWC 2019: 18th International Semantic Web Conference, Auckland, New Zealand, October 26--30, 2019, Proceedings, Part I 18}, pages 219--236. Springer.

\bibitem[{Intelligence(2024)}]{nova}
Amazon Artificial~General Intelligence. 2024.
\newblock The amazon nova family of models: Technical report and model card.
\newblock \url{https://assets.amazon.science/}.

\bibitem[{Jain et~al.()Jain, Vaidyanath, Iyer, Natarajan, Parthasarathy, Rajamani, and Sharma}]{Jigsaw}
Naman Jain, Skanda Vaidyanath, Arun Iyer, Nagarajan Natarajan, Suresh Parthasarathy, Sriram Rajamani, and Rahul Sharma.
\newblock Jigsaw: Large language models meet program synthesis.
\newblock In \emph{ICSE 2022}.

\bibitem[{Jiang et~al.(2024)Jiang, Liutianci@stu.jiangnan.edu.cn, and Lu}]{jiang-etal-2024-jn}
Yunfan Jiang, Liutianci@stu.jiangnan.edu.cn Liutianci@stu.jiangnan.edu.cn, and Heng-yang Lu. 2024.
\newblock \href {https://aclanthology.org/2024.sighan-1.14/} {{JN}-{NLP} at {SIGHAN}-2024 dim{ABSA} task: Extraction of sentiment intensity quadruples based on paraphrase generation}.
\newblock In \emph{Proceedings of the 10th SIGHAN Workshop on Chinese Language Processing (SIGHAN-10)}, pages 121--126, Bangkok, Thailand. Association for Computational Linguistics.

\bibitem[{Jo and Gebru(2020)}]{jo2020lessons}
Eun~Seo Jo and Timnit Gebru. 2020.
\newblock Lessons from archives: Strategies for collecting sociocultural data in machine learning.
\newblock In \emph{Proceedings of the 2020 conference on fairness, accountability, and transparency}, pages 306--316.

\bibitem[{Kenton and Toutanova(2019)}]{kenton2019bert}
Jacob Devlin Ming-Wei~Chang Kenton and Lee~Kristina Toutanova. 2019.
\newblock Bert: Pre-training of deep bidirectional transformers for language understanding.
\newblock In \emph{Proceedings of naacL-HLT}, volume~1, page~2. Minneapolis, Minnesota.

\bibitem[{Lauren{\c{c}}on et~al.(2022)Lauren{\c{c}}on, Saulnier, Wang, Akiki, Villanova~del Moral, Le~Scao, Von~Werra, Mou, Gonz{\'a}lez~Ponferrada, Nguyen et~al.}]{laurenccon2022bigscience}
Hugo Lauren{\c{c}}on, Lucile Saulnier, Thomas Wang, Christopher Akiki, Albert Villanova~del Moral, Teven Le~Scao, Leandro Von~Werra, Chenghao Mou, Eduardo Gonz{\'a}lez~Ponferrada, Huu Nguyen, et~al. 2022.
\newblock The bigscience roots corpus: A 1.6 tb composite multilingual dataset.
\newblock \emph{Advances in Neural Information Processing Systems}, 35:31809--31826.

\bibitem[{Lieber et~al.(2024)Lieber, Lenz, Bata, Cohen, Osin, Dalmedigos, Safahi, Meirom, Belinkov, Shalev-Shwartz et~al.}]{lieber2024jamba}
Opher Lieber, Barak Lenz, Hofit Bata, Gal Cohen, Jhonathan Osin, Itay Dalmedigos, Erez Safahi, Shaked Meirom, Yonatan Belinkov, Shai Shalev-Shwartz, et~al. 2024.
\newblock Jamba: A hybrid transformer-mamba language model.
\newblock \emph{arXiv preprint arXiv:2403.19887}.

\bibitem[{Liu(2019)}]{liu2019roberta}
Yinhan Liu. 2019.
\newblock Roberta: A robustly optimized bert pretraining approach.
\newblock \emph{arXiv preprint arXiv:1907.11692}, 364.

\bibitem[{Longpre et~al.(2024)Longpre, Yauney, Reif, Lee, Roberts, Zoph, Zhou, Wei, Robinson, Mimno, and Ippolito}]{longpre-etal-2024-pretrainers}
Shayne Longpre, Gregory Yauney, Emily Reif, Katherine Lee, Adam Roberts, Barret Zoph, Denny Zhou, Jason Wei, Kevin Robinson, David Mimno, and Daphne Ippolito. 2024.
\newblock \href {https://doi.org/10.18653/v1/2024.naacl-long.179} {A pretrainer`s guide to training data: Measuring the effects of data age, domain coverage, quality, {\&} toxicity}.
\newblock In \emph{Proceedings of the 2024 Conference of the North American Chapter of the Association for Computational Linguistics: Human Language Technologies (Volume 1: Long Papers)}, pages 3245--3276, Mexico City, Mexico. Association for Computational Linguistics.

\bibitem[{Luccioni and Viviano(2021)}]{luccioni2021s}
Alexandra~Sasha Luccioni and Joseph~D Viviano. 2021.
\newblock What's in the box? a preliminary analysis of undesirable content in the common crawl corpus.
\newblock \emph{arXiv preprint arXiv:2105.02732}.

\bibitem[{Lucy et~al.(2024)Lucy, Gururangan, Soldaini, Strubell, Bamman, Klein, and Dodge}]{lucy2024aboutme}
Li~Lucy, Suchin Gururangan, Luca Soldaini, Emma Strubell, David Bamman, Lauren~F Klein, and Jesse Dodge. 2024.
\newblock Aboutme: Using self-descriptions in webpages to document the effects of english pretraining data filters.
\newblock \emph{arXiv preprint arXiv:2401.06408}.

\bibitem[{Manghi(2023)}]{manghi2023miriam}
Paolo Manghi. 2023.
\newblock Miriam baglioni1, andrea mannocci1, gina pavone1, michele de bonis1 and.
\newblock pages 47--59.

\bibitem[{Mickus et~al.(2024)Mickus, Gr{\"o}nroos, Attieh, Boggia, de~Gibert, Ji, Loppi, Raganato, V{\'a}zquez, and Tiedemann}]{mickus2024mammoth}
Timothee Mickus, Stig-Arne Gr{\"o}nroos, Joseph Attieh, Michele Boggia, Ona de~Gibert, Shaoxiong Ji, Niki~Andreas Loppi, Alessandro Raganato, Ra{\'u}l V{\'a}zquez, and J{\"o}rg Tiedemann. 2024.
\newblock Mammoth: Massively multilingual modular open translation@ helsinki.
\newblock In \emph{Proceedings of the 18th Conference of the European Chapter of the Association for Computational Linguistics: System Demonstrations}, pages 127--136.

\bibitem[{Mitchell et~al.(2019)Mitchell, Wu, Zaldivar, Barnes, Vasserman, Hutchinson, Spitzer, Raji, and Gebru}]{mitchell2019model}
Margaret Mitchell, Simone Wu, Andrew Zaldivar, Parker Barnes, Lucy Vasserman, Ben Hutchinson, Elena Spitzer, Inioluwa~Deborah Raji, and Timnit Gebru. 2019.
\newblock Model cards for model reporting.
\newblock In \emph{Proceedings of the conference on fairness, accountability, and transparency}, pages 220--229.

\bibitem[{Penedo et~al.(2023)Penedo, Malartic, Hesslow, Cojocaru, Cappelli, Alobeidli, Pannier, Almazrouei, and Launay}]{penedo2023refinedweb}
Guilherme Penedo, Quentin Malartic, Daniel Hesslow, Ruxandra Cojocaru, Alessandro Cappelli, Hamza Alobeidli, Baptiste Pannier, Ebtesam Almazrouei, and Julien Launay. 2023.
\newblock The refinedweb dataset for falcon llm: outperforming curated corpora with web data, and web data only.
\newblock \emph{arXiv preprint arXiv:2306.01116}.

\bibitem[{Rae et~al.(2021)Rae, Borgeaud, Cai, Millican, Hoffmann, Song, Aslanides, Henderson, Ring, Young et~al.}]{rae2021scaling}
Jack~W Rae, Sebastian Borgeaud, Trevor Cai, Katie Millican, Jordan Hoffmann, Francis Song, John Aslanides, Sarah Henderson, Roman Ring, Susannah Young, et~al. 2021.
\newblock Scaling language models: Methods, analysis \& insights from training gopher.
\newblock \emph{arXiv preprint arXiv:2112.11446}.

\bibitem[{Raffel et~al.(2020)Raffel, Shazeer, Roberts, Lee, Narang, Matena, Zhou, Li, and Liu}]{raffel2020exploring}
Colin Raffel, Noam Shazeer, Adam Roberts, Katherine Lee, Sharan Narang, Michael Matena, Yanqi Zhou, Wei Li, and Peter~J Liu. 2020.
\newblock Exploring the limits of transfer learning with a unified text-to-text transformer.
\newblock \emph{Journal of machine learning research}, 21(140):1--67.

\bibitem[{Raffel et~al.(2024)Raffel, Agostinelli, and Chen}]{raffel-etal-2024-simultaneous}
Matthew Raffel, Victor Agostinelli, and Lizhong Chen. 2024.
\newblock \href {https://doi.org/10.18653/v1/2024.emnlp-main.1017} {Simultaneous masking, not prompting optimization: A paradigm shift in fine-tuning {LLM}s for simultaneous translation}.
\newblock In \emph{Proceedings of the 2024 Conference on Empirical Methods in Natural Language Processing}, pages 18302--18314, Miami, Florida, USA. Association for Computational Linguistics.

\bibitem[{Rajpurkar et~al.(2016)Rajpurkar, Zhang, Lopyrev, and Liang}]{rajpurkar2016squad}
Pranav Rajpurkar, Jian Zhang, Konstantin Lopyrev, and Percy Liang. 2016.
\newblock Squad: 100,000+ questions for machine comprehension of text.
\newblock In \emph{Proceedings of the 2016 Conference on Empirical Methods in Natural Language Processing}, pages 2383--2392.

\bibitem[{Research et~al.(2024)Research, An, Bae, Choi, Choi, Choi, Hong, Hwang, Jeon, Jo et~al.}]{research2024exaone}
LG~Research, Soyoung An, Kyunghoon Bae, Eunbi Choi, Kibong Choi, Stanley~Jungkyu Choi, Seokhee Hong, Junwon Hwang, Hyojin Jeon, Gerrard~Jeongwon Jo, et~al. 2024.
\newblock Exaone 3.5: Series of large language models for real-world use cases.
\newblock \emph{arXiv preprint arXiv:2412.04862}.

\bibitem[{Research(2023)}]{mosaic2023}
Mosaic Research. 2023.
\newblock Introducing mpt-7b: A new standard for open-source, commercially usable llms.
\newblock \url{https://www.databricks.com/blog/mpt-7b}.

\bibitem[{Soldaini et~al.(2024)Soldaini, Kinney, Bhagia, Schwenk, Atkinson, Authur, Bogin, Chandu, Dumas, Elazar, Hofmann, Jha, Kumar, Lucy, Lyu, Lambert, Magnusson, Morrison, Muennighoff, Naik, Nam, Peters, Ravichander, Richardson, Shen, Strubell, Subramani, Tafjord, Walsh, Zettlemoyer, Smith, Hajishirzi, Beltagy, Groeneveld, Dodge, and Lo}]{soldaini-etal-2024-dolma}
Luca Soldaini, Rodney Kinney, Akshita Bhagia, Dustin Schwenk, David Atkinson, Russell Authur, Ben Bogin, Khyathi Chandu, Jennifer Dumas, Yanai Elazar, Valentin Hofmann, Ananya Jha, Sachin Kumar, Li~Lucy, Xinxi Lyu, Nathan Lambert, Ian Magnusson, Jacob Morrison, Niklas Muennighoff, Aakanksha Naik, Crystal Nam, Matthew Peters, Abhilasha Ravichander, Kyle Richardson, Zejiang Shen, Emma Strubell, Nishant Subramani, Oyvind Tafjord, Evan Walsh, Luke Zettlemoyer, Noah Smith, Hannaneh Hajishirzi, Iz~Beltagy, Dirk Groeneveld, Jesse Dodge, and Kyle Lo. 2024.
\newblock \href {https://doi.org/10.18653/v1/2024.acl-long.840} {Dolma: an open corpus of three trillion tokens for language model pretraining research}.
\newblock In \emph{Proceedings of the 62nd Annual Meeting of the Association for Computational Linguistics (Volume 1: Long Papers)}, pages 15725--15788, Bangkok, Thailand. Association for Computational Linguistics.

\bibitem[{St et~al.(1989)St, Wold et~al.}]{st1989analysis}
Lars St, Svante Wold, et~al. 1989.
\newblock Analysis of variance (anova).
\newblock \emph{Chemometrics and intelligent laboratory systems}, 6(4):259--272.

\bibitem[{Stranisci et~al.(2023{\natexlab{a}})Stranisci, Bernasconi, Patti, Ferilli, Ceriani, and Damiano}]{stranisci2023world}
Marco~Antonio Stranisci, Eleonora Bernasconi, Viviana Patti, Stefano Ferilli, Miguel Ceriani, and Rossana Damiano. 2023{\natexlab{a}}.
\newblock The world literature knowledge graph.
\newblock In \emph{International Semantic Web Conference}, pages 435--452. Springer.

\bibitem[{Stranisci et~al.(2023{\natexlab{b}})Stranisci, Damiano, Mensa, Patti, Radicioni, and Caselli}]{stranisci2023wikibio}
Marco~Antonio Stranisci, Rossana Damiano, Enrico Mensa, Viviana Patti, Daniele Radicioni, and Tommaso Caselli. 2023{\natexlab{b}}.
\newblock Wikibio: a semantic resource for the intersectional analysis of biographical events.
\newblock In \emph{Proceedings of the 61st Annual Meeting of the Association for Computational Linguistics (Volume 1: Long Papers)}, pages 12370--12384.

\bibitem[{Su et~al.(2024)Su, Wu, Huang, Zhang, Wang, Hu, and Sha}]{su-etal-2024-refine}
Guixin Su, Mingmin Wu, Zhongqiang Huang, Yongcheng Zhang, Tongguan Wang, Yuxue Hu, and Ying Sha. 2024.
\newblock \href {https://doi.org/10.18653/v1/2024.findings-acl.191} {Refine, align, and aggregate: Multi-view linguistic features enhancement for aspect sentiment triplet extraction}.
\newblock In \emph{Findings of the Association for Computational Linguistics: ACL 2024}, pages 3212--3228, Bangkok, Thailand. Association for Computational Linguistics.

\bibitem[{Sun et~al.(2021)Sun, Wang, Feng, Ding, Pang, Shang, Liu, Chen, Zhao, Lu et~al.}]{sun2021ernie}
Yu~Sun, Shuohuan Wang, Shikun Feng, Siyu Ding, Chao Pang, Junyuan Shang, Jiaxiang Liu, Xuyi Chen, Yanbin Zhao, Yuxiang Lu, et~al. 2021.
\newblock Ernie 3.0: Large-scale knowledge enhanced pre-training for language understanding and generation.
\newblock \emph{arXiv preprint arXiv:2107.02137}.

\bibitem[{Taori et~al.(2023)Taori, Gulrajani, Zhang, Dubois, Li, Guestrin, Liang, and Hashimoto}]{alpaca}
Rohan Taori, Ishaan Gulrajani, Tianyi Zhang, Yann Dubois, Xuechen Li, Carlos Guestrin, Percy Liang, and Tatsunori~B. Hashimoto. 2023.
\newblock Stanford alpaca: An instruction-following llama model.
\newblock \url{https://github.com/tatsu-lab/stanford_alpaca}.

\bibitem[{Team et~al.(2024{\natexlab{a}})Team, Mesnard, Hardin, Dadashi, Bhupatiraju, Pathak, Sifre, Rivi{\`e}re, Kale, Love et~al.}]{team2024gemma}
Gemma Team, Thomas Mesnard, Cassidy Hardin, Robert Dadashi, Surya Bhupatiraju, Shreya Pathak, Laurent Sifre, Morgane Rivi{\`e}re, Mihir~Sanjay Kale, Juliette Love, et~al. 2024{\natexlab{a}}.
\newblock Gemma: Open models based on gemini research and technology.
\newblock \emph{arXiv preprint arXiv:2403.08295}.

\bibitem[{team(2024)}]{grok}
Grok team. 2024.
\newblock Grok-2 beta release.
\newblock \url{https://x.ai/blog/grok-2}.

\bibitem[{Team et~al.(2024{\natexlab{b}})Team, Ormazabal, Zheng, d'Autume, Yogatama, Fu, Ong, Chen, Lamprecht, Pham et~al.}]{team2024reka}
Reka Team, Aitor Ormazabal, Che Zheng, Cyprien de~Masson d'Autume, Dani Yogatama, Deyu Fu, Donovan Ong, Eric Chen, Eugenie Lamprecht, Hai Pham, et~al. 2024{\natexlab{b}}.
\newblock Reka core, flash, and edge: A series of powerful multimodal language models.
\newblock \emph{arXiv preprint arXiv:2404.12387}.

\bibitem[{Tedeschi et~al.(2024)Tedeschi, Friedrich, Schramowski, Kersting, Navigli, Nguyen, and Li}]{tedeschi2024alert}
Simone Tedeschi, Felix Friedrich, Patrick Schramowski, Kristian Kersting, Roberto Navigli, Huu Nguyen, and Bo~Li. 2024.
\newblock Alert: A comprehensive benchmark for assessing large language models' safety through red teaming.
\newblock \emph{arXiv preprint arXiv:2404.08676}.

\bibitem[{Thoppilan et~al.(2022)Thoppilan, De~Freitas, Hall, Shazeer, Kulshreshtha, Cheng, Jin, Bos, Baker, Du et~al.}]{thoppilan2022lamda}
Romal Thoppilan, Daniel De~Freitas, Jamie Hall, Noam Shazeer, Apoorv Kulshreshtha, Heng-Tze Cheng, Alicia Jin, Taylor Bos, Leslie Baker, Yu~Du, et~al. 2022.
\newblock Lamda: Language models for dialog applications.
\newblock \emph{arXiv preprint arXiv:2201.08239}.

\bibitem[{Touvron et~al.(2023)Touvron, Martin, Stone, Albert, Almahairi, Babaei, Bashlykov, Batra, Bhargava, Bhosale et~al.}]{touvron2023llama}
Hugo Touvron, Louis Martin, Kevin Stone, Peter Albert, Amjad Almahairi, Yasmine Babaei, Nikolay Bashlykov, Soumya Batra, Prajjwal Bhargava, Shruti Bhosale, et~al. 2023.
\newblock Llama 2: Open foundation and fine-tuned chat models.
\newblock \emph{arXiv preprint arXiv:2307.09288}.

\bibitem[{Tunstall et~al.(2023)Tunstall, Beeching, Lambert, Rajani, Rasul, Belkada, Huang, von Werra, Fourrier, Habib et~al.}]{tunstall2023zephyr}
Lewis Tunstall, Edward Beeching, Nathan Lambert, Nazneen Rajani, Kashif Rasul, Younes Belkada, Shengyi Huang, Leandro von Werra, Cl{\'e}mentine Fourrier, Nathan Habib, et~al. 2023.
\newblock Zephyr: Direct distillation of lm alignment.
\newblock \emph{arXiv preprint arXiv:2310.16944}.

\bibitem[{Wake et~al.(2024)Wake, Wang, Chen, Lv, Li, Huang, Cai, Zheng, Cooper, Dai et~al.}]{wake2024yi}
Alan Wake, Albert Wang, Bei Chen, CX~Lv, Chao Li, Chengen Huang, Chenglin Cai, Chujie Zheng, Daniel Cooper, Ethan Dai, et~al. 2024.
\newblock Yi-lightning technical report.
\newblock \emph{arXiv preprint arXiv:2412.01253}.

\bibitem[{Wang et~al.(2019)Wang, Pruksachatkun, Nangia, Singh, Michael, Hill, Levy, and Bowman}]{wang2019superglue}
Alex Wang, Yada Pruksachatkun, Nikita Nangia, Amanpreet Singh, Julian Michael, Felix Hill, Omer Levy, and Samuel Bowman. 2019.
\newblock Superglue: A stickier benchmark for general-purpose language understanding systems.
\newblock \emph{Advances in neural information processing systems}, 32.

\bibitem[{Wang et~al.(2024)Wang, Ma, Zhang, Ni, Chandra, Guo, Ren, Arulraj, He, Jiang et~al.}]{wang2024mmlu}
Yubo Wang, Xueguang Ma, Ge~Zhang, Yuansheng Ni, Abhranil Chandra, Shiguang Guo, Weiming Ren, Aaran Arulraj, Xuan He, Ziyan Jiang, et~al. 2024.
\newblock Mmlu-pro: A more robust and challenging multi-task language understanding benchmark.
\newblock \emph{arXiv preprint arXiv:2406.01574}.

\bibitem[{Weber et~al.(2024)Weber, Fu, Anthony, Oren, Adams, Alexandrov, Lyu, Nguyen, Yao, Adams et~al.}]{weber2024redpajama}
Maurice Weber, Daniel Fu, Quentin Anthony, Yonatan Oren, Shane Adams, Anton Alexandrov, Xiaozhong Lyu, Huu Nguyen, Xiaozhe Yao, Virginia Adams, et~al. 2024.
\newblock Redpajama: an open dataset for training large language models.
\newblock \emph{arXiv preprint arXiv:2411.12372}.

\bibitem[{Xie et~al.(2023)Xie, Santurkar, Ma, and Liang}]{xie2023data}
Sang~Michael Xie, Shibani Santurkar, Tengyu Ma, and Percy~S Liang. 2023.
\newblock Data selection for language models via importance resampling.
\newblock \emph{Advances in Neural Information Processing Systems}, 36:34201--34227.

\bibitem[{Xu et~al.(2021)Xu, Pathak, Wallace, Gururangan, Sap, and Klein}]{xu2021detoxifying}
Albert Xu, Eshaan Pathak, Eric Wallace, Suchin Gururangan, Maarten Sap, and Dan Klein. 2021.
\newblock Detoxifying language models risks marginalizing minority voices.
\newblock In \emph{Proceedings of the 2021 Conference of the North American Chapter of the Association for Computational Linguistics: Human Language Technologies}, pages 2390--2397.

\bibitem[{Yang(2019)}]{yang2019xlnet}
Zhilin Yang. 2019.
\newblock Xlnet: Generalized autoregressive pretraining for language understanding.
\newblock \emph{arXiv preprint arXiv:1906.08237}.

\bibitem[{Yasunaga et~al.(2022)Yasunaga, Leskovec, and Liang}]{yasunaga2022linkbert}
Michihiro Yasunaga, Jure Leskovec, and Percy Liang. 2022.
\newblock Linkbert: Pretraining language models with document links.
\newblock In \emph{Proceedings of the 60th Annual Meeting of the Association for Computational Linguistics (Volume 1: Long Papers)}, pages 8003--8016.

\bibitem[{Zhong et~al.(2022)Zhong, Ding, Zhan, Qiao, Wen, Shen, Liu, Yu, Du, Chen et~al.}]{zhong2022toward}
Qihuang Zhong, Liang Ding, Yibing Zhan, Yu~Qiao, Yonggang Wen, Li~Shen, Juhua Liu, Baosheng Yu, Bo~Du, Yixin Chen, et~al. 2022.
\newblock Toward efficient language model pretraining and downstream adaptation via self-evolution: A case study on superglue.
\newblock \emph{arXiv preprint arXiv:2212.01853}.

\bibitem[{Zhu et~al.(2015)Zhu, Kiros, Zemel, Salakhutdinov, Urtasun, Torralba, and Fidler}]{7410368}
Yukun Zhu, Ryan Kiros, Rich Zemel, Ruslan Salakhutdinov, Raquel Urtasun, Antonio Torralba, and Sanja Fidler. 2015.
\newblock \href {https://doi.org/10.1109/ICCV.2015.11} {Aligning books and movies: Towards story-like visual explanations by watching movies and reading books}.
\newblock In \emph{2015 IEEE International Conference on Computer Vision (ICCV)}, pages 19--27.

\end{thebibliography}
